\pdfoutput=1
\documentclass[11pt]{article}

\usepackage[final]{acl}

\usepackage{times}
\usepackage{latexsym}

\usepackage[T1]{fontenc}
\usepackage[utf8]{inputenc}

\usepackage{microtype}

\usepackage{inconsolata}

\usepackage{graphicx}

\usepackage{bm}
\usepackage{booktabs}
\usepackage{amsmath}
\usepackage{graphicx}
\usepackage{multirow}
\usepackage{hyperref}
\usepackage{cleveref}
\usepackage{url}
\usepackage{algorithm}
\usepackage{algorithmicx}
\usepackage[noend]{algpseudocode}
\usepackage{enumitem}
\usepackage{soul}

\usepackage{CJKutf8}
\newcommand{\zh}[1]{\begin{CJK*}{UTF8}{gkai}#1\end{CJK*}}
\usepackage{xcolor}
\definecolor{hzw}{RGB}{223, 97, 76}

\usepackage{listings}
\colorlet{punct}{red!60!black}
\definecolor{background}{HTML}{EEEEEE}
\definecolor{delim}{RGB}{20,105,176}
\colorlet{numb}{magenta!60!black}
\lstdefinelanguage{json}{
    basicstyle=\normalfont\ttfamily,
    numbers=none,
    numberstyle=\scriptsize,
    stepnumber=1,
    numbersep=8pt,
    tabsize=1,
    escapeinside={(*@}{@*)},
    showstringspaces=false,
    breaklines=true,
    frame=none,
    breakatwhitespace=true,
    backgroundcolor=\color{background},
    literate=
     *{0}{{{\color{numb}0}}}{1}
      {1}{{{\color{numb}1}}}{1}
      {2}{{{\color{numb}2}}}{1}
      {3}{{{\color{numb}3}}}{1}
      {4}{{{\color{numb}4}}}{1}
      {5}{{{\color{numb}5}}}{1}
      {6}{{{\color{numb}6}}}{1}
      {7}{{{\color{numb}7}}}{1}
      {8}{{{\color{numb}8}}}{1}
      {9}{{{\color{numb}9}}}{1}
      {:}{{{\color{punct}{:}}}}{1}
      {,}{{{\color{punct}{,}}}}{1}
      {\{}{{{\color{delim}{\{}}}}{1}
      {\}}{{{\color{delim}{\}}}}}{1}
      {[}{{{\color{delim}{[}}}}{1}
      {]}{{{\color{delim}{]}}}}{1},
}

\usepackage{arydshln}

\title{Encouraging Divergent Thinking in Large Language Models \\ through Multi-Agent Debate}

\author{
Tian Liang$^1$$^,$$^3$\thanks{\ \ Contributed equally. Work was done when Tian and Zhiwei were interning at Tencent AI Lab.}  \quad Zhiwei He$^2$\footnotemark[1] \quad Wenxiang Jiao$^{3}$\footnotemark[1] \quad  Xing Wang$^{3}$\footnotemark[2] \quad Yan Wang$^{3}$ \\ 
\bf Rui Wang$^2$ \quad Yujiu Yang$^{1}$\thanks{\ \ Xing and Yujiu are co‐corresponding authors.} \quad Shuming Shi$^3$ \quad  Zhaopeng Tu$^3$\\
$^{1}$Tsinghua University \ \ \ $^{2}$Shanghai Jiao Tong University \ \ \  $^{3}$Tencent AI Lab \\
$^1$\texttt{\{liangt21@mails,yang.yujiu@sz\}.tsinghua.edu.cn} \\
$^2$\texttt{zwhe.cs@sjtu.edu.cn} ~~~~ $^3$\texttt{\{joelwxjiao,brightxwang,zptu\}@tencent.com}
}

\begin{document}
\maketitle

\begin{abstract}
Modern large language models~(LLMs) like ChatGPT have shown remarkable performance on general language tasks but still struggle on complex reasoning tasks, which drives the research on cognitive behaviors of LLMs to explore human-like problem-solving strategies. 
Along this direction, one representative strategy is self-reflection, which asks an LLM to refine the solution with the feedback generated by itself iteratively. 
However, our study shows that such reflection-style methods suffer from the \textit{Degeneration-of-Thought~(DoT)} problem: once the LLM has established confidence in its solutions, it is unable to generate novel thoughts later through reflection even if its initial stance is incorrect.
To address the DoT problem, we propose a \textbf{Multi-Agent Debate~(MAD)} framework, in which multiple agents express their arguments in the state of ``tit for tat'' and a judge manages the debate process to obtain a final solution. 
Clearly, our MAD framework encourages divergent thinking in LLMs which would be helpful for tasks that require deep levels of contemplation.
Experiment results on two challenging datasets, commonsense machine translation and counter-intuitive arithmetic reasoning, demonstrate the effectiveness of our MAD framework. 
Extensive analyses suggest that the adaptive break of debate and the modest level of ``tit for tat'' state are required for MAD to obtain good performance. Moreover, we find that LLMs might not be a fair judge if different LLMs are used for agents. 
Code is available at \url{https://github.com/Skytliang/Multi-Agents-Debate}.

\end{abstract}
\section{Introduction}
\label{sec:introduction}

\begin{figure}[t]
    \centering
    \includegraphics[width=0.8\linewidth]{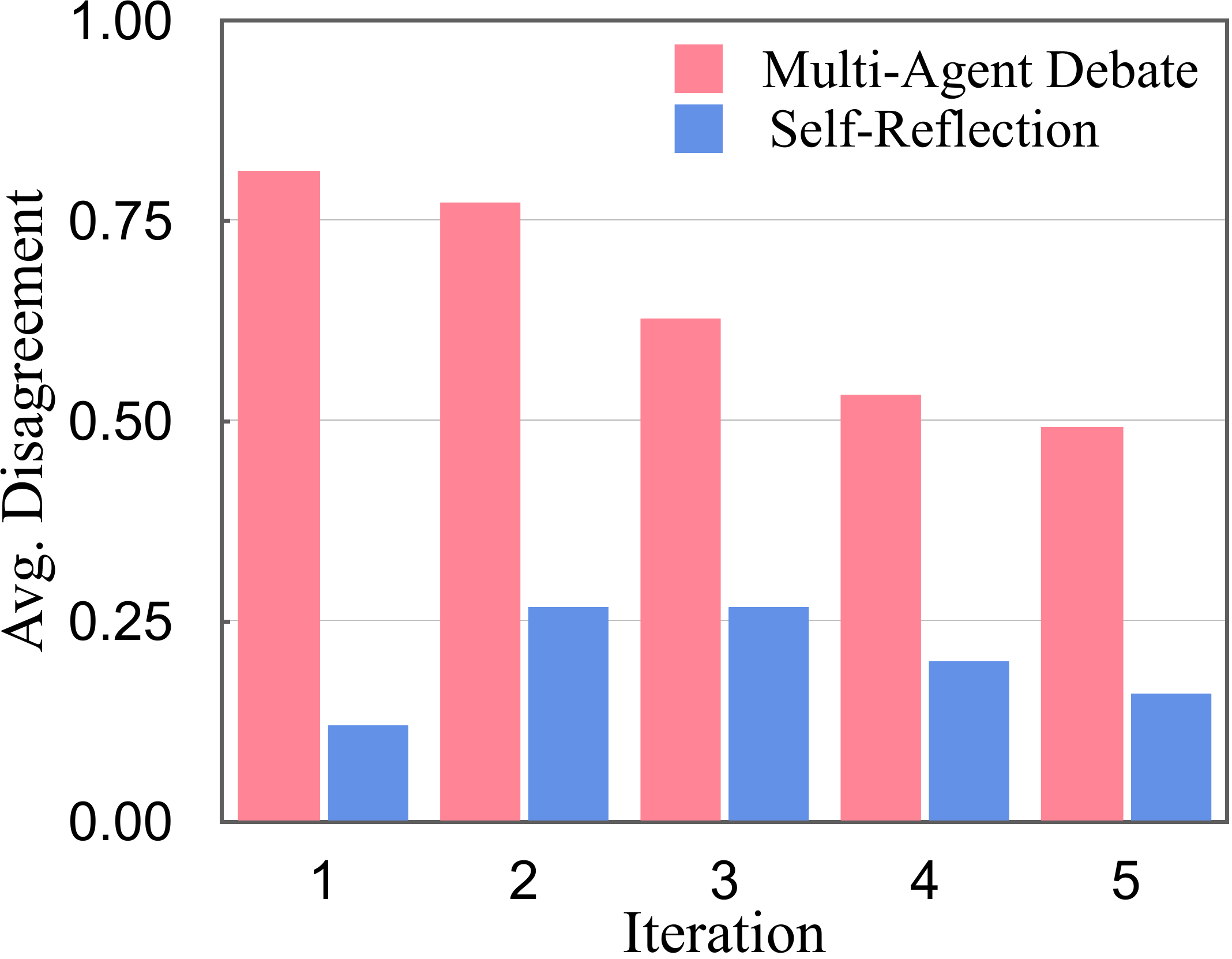}
    \vspace{-7pt}
    \caption{Disagreement between two adjacent iterations with respect to the iteration of debate/self-reflection.}
    \label{fig:intro}
    \vspace{-15pt}
\end{figure}

Large language models~(LLMs) have shown remarkable performance on general language tasks~\citep{jiao2023chatgpt,wu2023chatgpt4gec,bang2023multitask} but still struggle on complex reasoning tasks~\citep{zhu2022core, gou2023critic}, which drives the research on cognitive behaviors of LLMs to explore human-like problem-solving strategies. 
In particular, \textit{self-reflection}~\citep{madaan2023selfrefine,shinn2023reflexion}, a concept that usually refers to the process of introspection and examination of a person's own thoughts, has been explored to solve intricate tasks that could be challenging for a zero-shot generation or even chain-of-thought~(CoT) prompting~\citep{wei2022CoT}.
Specifically, self-reflection involves an iterative refinement process such that the LLM generates a new answer based on the answers and feedback in previous iterations and then provides feedback for the new answer. 
While self-reflection can be effective in creating better answers, it is highly dependent on the self-evaluation capabilities of LLMs, which are not formally guaranteed~\citep{shinn2023reflexion}.  

In this work, we focus on the \textbf{Degeneration-of-Thought}~(DoT) problem in self-reflection, which is proposed and defined by us for the first time. Formally, DoT describes the following scenario: 
\begin{quote}
    \it
    Once the LLM-based agent has established confidence in its answers, it is unable to generate novel thoughts later through self-reflection even if the initial stance is incorrect. 
\end{quote}

To demonstrate this problem, we force the agents to engage in a debate or self-reflection for 5 rounds before reaching an answer. Next, we manually determine the disagreement as 1 and agreement as 0 between two adjacent iterations. We define the average disagreement in iteration $i$ as the percentage of opposition occurring between two debaters across multiple debates (or self-confliction in self-reflection).
We show the trends in Figure~\ref{fig:intro}.
The low disagreement of self-reflection suggests that the LLM sticks to the incorrect answers predicted by CoT and is unable to engage in meaningful self-reflection.

There are various factors~\citep{bortolotti2011does,keestra2017metacognition} that could result in DoT, and we outline three here:
(1) \textit{Bias and Distorted Perception}. Self-perception can be influenced by biases, preconceived notions, and distorted thinking patterns, which can be learned from the massive amount of data during pretraining. If an LLM's self-reflection is clouded by such biases or distorted thinking, it can lead to inaccurate conclusions instinctively.
(2) \textit{Rigidity and Resistance to Change}. Self-reflection often involves challenging one's beliefs, assumptions, and behaviors. If an LLM is resistant to change or holds rigid beliefs, it may struggle to engage in meaningful self-reflection that leads to better answers.
(3) \textit{Limited External Feedback}. Self-reflection is primarily an internal process, but external feedback can provide valuable perspectives and insights. Without considering external feedback, an LLM may miss important blind spots or alternative viewpoints that can enrich its self-reflection.

To address the DoT issue, we leverage another fundamental characteristic of human problem-solving, i.e., debate, to encourage divergent thinking in LLMs.
Specifically, we propose the \textbf{MAD} framework, short for \textbf{M}ulti-\textbf{A}gent \textbf{D}ebate, where two agents express their own arguments in the state of ``tit for tat'' and a judge monitors and manages the debate process to obtain a final solution.
The nature of MAD determines that (1) The distorted thinking of one agent can be corrected by the others; (2) The resistance to change of one agent will be complemented by the others; and (3) each agent can obtain external feedback from the others. 
Therefore, MAD is less susceptible to the factors of DoT, and can explore divergent chain-of-thoughts to achieve accurate answers.

We conducted experiments on both natural language generation and understanding through two challenging tasks, namely, Commonsense Machine Translation~(Common~MT) and Counter-Intuitive Arithmetic Reasoning~(Counter-Intuitive~AR).
The common characteristic of the two tasks is that our instincts are mostly incorrect based on only the superficial expressions of the questions, and deeper levels of contemplation are required for better answers.
Experimental results demonstrate that our MAD framework outperforms the baseline methods, especially, \texttt{GPT-3.5-Turbo} with MAD can surpass the performance of \texttt{GPT-4} on Common~MT. 

The contributions of this work are summarized as follows:
\begin{itemize}[leftmargin=10pt]
    \item We propose and define the Degeneration-of-Thought (DoT) problem in self-reflection, and address it by proposing the Multi-Agent Debate~(MAD) framework to explore divergent chain-of-thoughts.
    
    \item We demonstrate the effectiveness of MAD on two challenging tasks, and find that \texttt{GPT-3.5-Turbo} with MAD can even surpass \texttt{GPT-4} on the Common~MT dataset.

    \item Experimental results show that the adaptive break strategy and the modest level of ``tit for tat'' state are required for performance improvement. In addition, we find that the llm-based judge shows a preference to the side with the same LLM as the backbone.
\end{itemize}

\section{Multi-Agent Debate Framework}
\label{sec:method}

Figure~\ref{fig:method} illustrates the general framework of MAD, where two debaters and a judge are involved in a debate to resolve a math problem while self-reflection descends into the trap of DoT. Generally, our MAD framework is composed of three components which are elaborated as follows:

\begin{figure*}[t!]
    \centering
    \includegraphics[width=\textwidth]{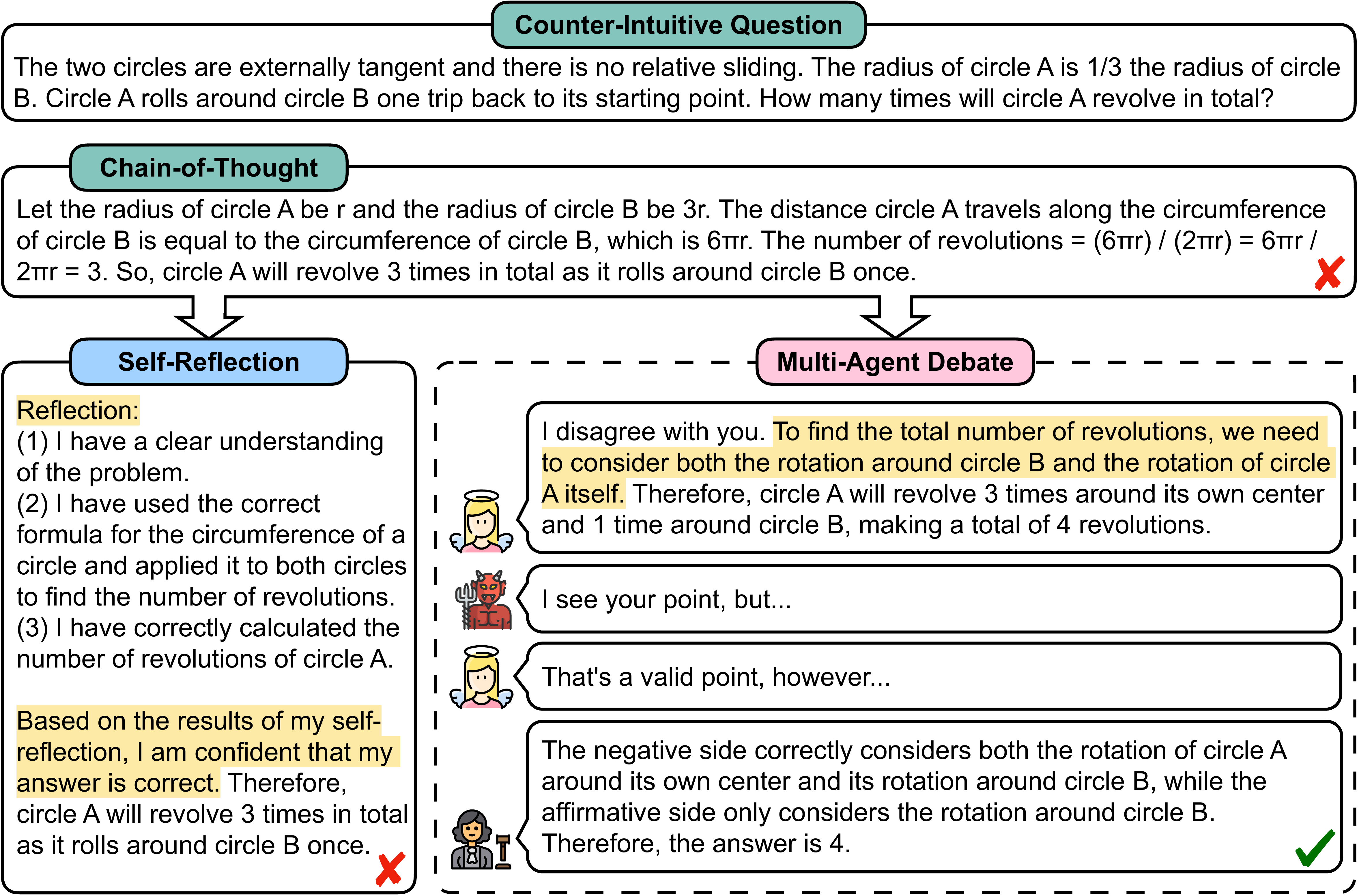}
    \vspace{-20pt}
    \caption{Framework of Multi-Agent Debate. Here we designate the devil (\begin{minipage}[b]{0.06\columnwidth}
    \centering
    \raisebox{-.3\linewidth}{\includegraphics[width=\linewidth]{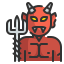}}
\end{minipage}) as the affirmative side while the angel (\begin{minipage}[b]{0.06\columnwidth}
    \centering
    \raisebox{-.3\linewidth}{\includegraphics[width=\linewidth]{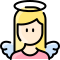}} 
\end{minipage}) as the negative side. We want the angel to correct the devil's mistakes.
}
    \label{fig:method}
    \vspace{-10pt}
\end{figure*}

\paragraph{Meta Prompts.}
We use meta prompts to introduce the topic to be solved, the number of debaters, the iteration limit, and other requirements. An exampe of meta prompts for the arithmetic reasoning task in Figure~\ref{fig:method} is:
\begin{quote}
    \it
    \setlength{\leftskip}{-1em}
     You are a debater. Hello and welcome to the debate competition. \ul{It's not necessary to fully agree with each other's perspectives, as our objective is to find the correct answer.} The debate topic is stated as follows: <debate topic>.
\end{quote}
As seen, we require the agents to ``tit for tat'' (e.g. contents underlined in meta prompts above) so as to create an atmosphere of debate. 

\paragraph{Debaters.}
There are $N$ debaters $D = \{D_i\}_{i=1}^N$ involved in the framework. 
In each debate iteration, the debaters $D_i$ speak one by one in a fixed order and express their arguments based on the previous debate history $H$, i.e., $D_i(H) = h$.  An example of a debater prompt appears below:

\begin{itemize}[leftmargin=1em]
    \item {Prompt for Affirmative Debater (\begin{minipage}[b]{0.06\columnwidth}
    \centering
    \raisebox{-.3\height}{\includegraphics[width=\linewidth]{imgs/emojis/devil.png}}
\end{minipage})}

\begin{quote}
    \it
    \setlength{\leftskip}{-1em}
     You are affirmative side. Please express your viewpoints.
\end{quote}

    \item {Prompt for Negative Debater (\begin{minipage}[b]{0.06\columnwidth}
    \centering
    \raisebox{-.3\height}{\includegraphics[width=\linewidth]{imgs/emojis/angel.png}}
\end{minipage})}

\begin{quote}
    \it
    \setlength{\leftskip}{-1em}
     You are negative side. You disagree with the affirmative side's points. Provide your reasons and answer. 
\end{quote}
\end{itemize}

\paragraph{Judge.}
We also design a judge $J$ to manage and monitor the whole debate process. 
The judge contains two different modes: (a) \textit{Discrinative Mode}, in which the judge $J$ decides whether the correct solution can be obtained after all the debaters finish their arguments in the current iteration:
\begin{equation}
    J_d(H)= \begin{cases}
        \texttt{True},~~ & \textrm{solution obtained} \\
        \texttt{False},~~ & \textrm{otherwise}
    \end{cases}
\end{equation}
If it is \texttt{True}, the debate is over. Otherwise, the debate continues. (b) \textit{Extractive Mode}, in which the judge $J$ needs to extract the final solution based on the whole debate history: $J_e(H) = a$, since no correct solution is identified within the iteration limit of debate. An example of a judge prompt (\begin{minipage}[b]{0.06\columnwidth}
    \centering
    \raisebox{-.3\height}{\includegraphics[width=\linewidth]{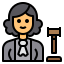}}
\end{minipage}) appears below:
\begin{quote}
    \it
    \setlength{\leftskip}{-1em}
    You are a moderator. There will be two debaters involved in a debate competition. They will present their answers and discuss their perspectives on the <debate topic>. At the end of each round, you will evaluate both sides' answers and decide which one is correct.
\end{quote}

\section{Experiment}

\subsection{Challenging Testbeds}

We conduct experiments on two challenging tasks, namely, commonsense machine translation (i.e., Common~MT), and counter-intuitive arithmetic reasoning (i.e., Counter-Intuitive AR), which require deep levels of contemplation for LLMs. Please refer to Appendix~\ref{app:testbeds} for more details.

\paragraph{Commonsense Machine Translation}
The Common~MT dataset is composed of Chinese$\Rightarrow$English translation examples~\citep{he-etal-2020-box}, which are used to examine three types of ambiguity resolution abilities of translation models, covering lexical
and contextless/contextual syntactic ambiguity. Within the challenging part of Common~MT, the authentic translation of each source sentence requires a proper understanding of common sense knowledge. While these ambiguous sentences might appear to have a straightforward translation, such a literal interpretation is erroneous. Failure to address such ambiguities may result in inaccurate translations.

\paragraph{Counter-Intuitive Arithmetic Reasoning}
Previous studies on thinking hierarchy~\citep{kahneman2011thinking} suggest that we humans have a fast and intuitive system and a slow and logical system, and tend to run the lower level system before the higher level one. Inspired by this, we created a more challenging dataset named Counter-Intuitive Arithmetic Reasoning (CIAR) to evaluate the reasoning abilities of LLMs at deep levels. 
Our Counter-Intuitive AR dataset contains 200 questions collected from elicitation 
questions~\citep{kong2022eliciting}\footnote{https://elicitation.info/questionnaire/1/}, 
web data\footnote{https://www.geeksforgeeks.org/puzzles/} and additional manual derivatives of these questions.
Compared to the commonly-used datasets, e.g., MultiArith~\citep{roy2015solving}, GSM8K~\citep{cobbe2021training},  our dataset presents two distinct challenges:
\begin{itemize}[leftmargin=10pt]
    \item \textit{Resistance to Intuition.} 
    The questions are embedded in hidden traps designed to elicit intuitive and appealing answers that are often incorrect. This feature evaluates the abilities of LLMs to resist the traps of superficial expressions.
    \item \textit{Multi-Step Reasoning.} 
    Each correct answer within the dataset requires a rigorous multi-step reasoning process, thereby evaluating the capacity of LLMs to engage in complex decision-making and problem-solving.
\end{itemize}

\subsection{Setups}
\paragraph{Input Format.}
Our experiments are performed in zero-shot instructions (setting temperature to 0).
For all used datasets, we use a unified prompt to make LLMs give explanations and answers.
We present the inputs to agents through $<$debate topic$>$ as mentioned in Section~\ref{sec:method}. For example, if we want to translate ``\zh{吃掉敌人一个师}'' from Chinese to English, we will set the $<$debate topic$>$ as \textit{``What is the correct English translation of the following Chinese text: \zh{吃掉敌人一个师}''}. For QA task, we employ the same prompt except set the $<$debate topic$>$ to the arithmetic question.

\paragraph{Backbone Models.}
In this work, we mainly use three agents in our MAD framework, including two debaters (i.e., affirmative and negative) and a judge. 
We assess two open-source (i.e., \texttt{vicuna-7b-v1.5-16k}\footnote{https://huggingface.co/lmsys/vicuna-7b-v1.5-16k} and \texttt{vicuna-13b-v1.5-16k}\footnote{https://huggingface.co/lmsys/vicuna-13b-v1.5-16k}) and two api-based LLMs (i.e., \texttt{GPT-3.5-Turbo-0301} and \texttt{GPT-4-0314}).

\begin{table*}[t]
\centering
\resizebox{\linewidth}{!}{
\begin{tabular}{l ccc ccc ccc}
\toprule
\multirow{2}{*}{\bf Method} & \multicolumn{3}{c}{\bf \normalsize Lexical} & \multicolumn{3}{c}{\bf \normalsize Contextless} & \multicolumn{3}{c}{\bf \normalsize Contextual} \\
\cmidrule(lr){2-4} \cmidrule(lr){5-7} \cmidrule(lr){8-10}
 & \small COMET & \small BLEURT & \small HUMAN & \small COMET & \small BLEURT & \small HUMAN & \small COMET & \small BLEURT & \small HUMAN\\
\midrule
\bf GPT-4  & \textbf{82.0} & 70.1 & 3.41 & 84.7 & 73.6 & 3.63 & 85.0 & 73.7 & 3.65 \\
\midrule
\bf \texttt{Turbo}  & 80.3 & 68.2 & 3.14 & 84.0 & 72.9 & 3.43 & 84.9 & 73.4 & 3.57 \\
~+ \textbf{Rerank}  & 80.9 & 68.6 & 3.16 & 84.5 & 73.2 & 3.46 & \textbf{85.3} & 73.9 & 3.58 \\
~+ \textbf{MAPS}  & 81.9 & 70.1 & 3.43 & 84.2 & 73.5 & 3.45 & 85.2 & \textbf{74.0} & 3.56 \\
~+ \textbf{\small Self-Reflect}  & 81.0 & 69.1 & 3.43 & 83.6 & 72.2 & 3.46 & 84.9 & 73.5 & 3.63 \\
\hdashline
~+ \textbf{MAD}  & \textbf{82.0} & \textbf{70.9} & \textbf{3.78} & \textbf{84.8} & \textbf{73.7} & \textbf{3.67} & \textbf{85.3} & \textbf{74.0} & \textbf{3.67} \\
\midrule
\bf \texttt{Vicuna-7b}  & 74.9 & 62.0 & 2.55  & 78.3 & 64.6 &  2.53  & 80.2 & 68.2 &  3.23  \\
~+ \textbf{MAD}  & 75.6 & 62.6 & 2.67  & 78.6 & 66.0 &  2.69  & 81.8 & 69.9 &  3.27  \\
\bf \texttt{Vicuna-13b}  & 76.6 & 63.7 & 2.81  & 77.6 & 66.8 &  3.04  & 82.2 & 70.0 &  3.37  \\
~+ \textbf{MAD}  & 77.2 & 65.1 & 2.96  & 80.1 & 67.3 & 3.11   & 82.6 & 70.9 &  3.45  \\
\bottomrule
\end{tabular}
}
\vspace{-5pt}
\caption{Translation performance on Common~MT. Note that Rerank and MAPS use the external quality estimation tool to select the best translation from multiple translation candidates. HUMAN: direct assessment of translation quality from human evaluators on a scale ranging from 1 to 5.}
\label{tab:common-mt}
\vspace{-10pt}
\end{table*}

\paragraph{Compared Methods.}
Generally, we compare our MAD framework with baseline models and Self-Reflect on both tasks. We also include other baseline methods individually, namely, Rerank and MAPS for Common~MT, CoT and Self-Consistency for Counter-Intuitive~AR. Below elaborates the details of them:
\begin{itemize}[leftmargin=10pt]
    \item \textbf{Self-Reflect}~\citep{shinn2023reflexion}: This approach requires the LLM to refine its translation until it deems the current output satisfactory.
    
    \item \textbf{Rerank}~\citep{he2023exploring}: We sample the translations from the LLM for four times, from which we select the best candidate based on a quality estimation (QE) HUMANr\footnote{We use \texttt{wmt21-comet-qe-da} as the QE HUMANr.}.
    This approach can be seen as analogous to self-consistency~\citep{wang2022self}, where the majority voting is replaced by an external QE HUMANr. 
    \item \textbf{MAPS}~\citep{he2023exploring}: This method enables LLMs to mimic the human translation process: analyze before translate, which can be viewed as a chain-of-thought method applied to translation. 
    
    \item \textbf{CoT}~\citep{kojimalarge}: This approach concatenates a trigger sentence ``Let’s think step by step''  to the test question.
    \item \textbf{Self-Consistency}~\citep{wang2022self}: This method samples multiple responses and determines the final answer through a majority vote. 
   
\end{itemize}

All agents in our experimental setup, such as debaters and judge, are large language models. Here, we implement the methods on top of \texttt{GPT-3.5-Turbo} and \texttt{Vicuna} models.

\paragraph{Evaluation Metrics.}
For Counter-Intuitive AR, we report the accuracy~(ACC) of predictions.
For Common~MT, we adopt automatic metrics like COMET\footnote{\texttt{\url{https://github.com/Unbabel/COMET/}, Unbabel/wmt22-comet-da}} and BLEURT\footnote{\texttt{\url{https://github.com/google-research/bleurt}, BLEURT-20}}, which are widely adopted evaluation metrics for LLM-based translation literature~\citep{he2023exploring,gpt-mt-2023,garcia2023unreasonable,pilault2023interactive}. 
In addition, we also employ professional human translators to directly assess the translation results, measuring translation quality on a scale ranging from 1 to 5.

\subsection{Results on Common~MT}
\label{subsec_mt}

\paragraph{Results.}
In Common~MT test set, we focus more on the translation accuracy of specific words and whether they conform to common sense. However, such minor variations at token level are difficult to reflect on automatic metrics. We therefore provide human HUMAN to evaluate these methods more accurately.
Table~\ref{tab:common-mt} presents the experimental results. MAPS and Self-Reflec achieve improvements over baseline \texttt{GPT-3.5-Turbo}. Remarkably, our proposed MAD, by utilizing GPT-3.5 as the backbone model, has demonstrated significant advancements over GPT-4 across both automatic and human evaluation metrics.

\begin{table}[!tb]
\centering
\resizebox{\columnwidth}{!}{
\begin{tabular}{l p{5cm}}
\toprule
\bf Source & \zh{\underline{吃掉}敌人一个师。}\\
\bf Correct Ref. & \textcolor{blue}{Destroy} a division of the enemy.\\
\bf Incorrect Ref. & \textcolor{red}{Eat up} an enemy division.\\
\midrule
\bf \texttt{GPT-4} & \textcolor{red}{Eat up} an enemy division. \\
\bf \texttt{GPT-3.5-Turbo} & \textcolor{red}{Eat up} an enemy division. \\
~~+ \textbf{Self-Reflect}  & \textcolor{red}{Eat up} an enemy division. \\
\hdashline
~~+ \textbf{MAD} & \textcolor{blue}{Eliminate} an enemy division.  \\
\bottomrule
\end{tabular}
}
\vspace{-5pt}
\caption{Example translations generated by different methods. Best viewed in color.}
\label{tab:mt-case}
\vspace{-10pt}
\end{table}

\paragraph{Case Study.} 
Table~\ref{tab:mt-case} shows example translations generated by baseline \texttt{GPT-3.5-Turbo} and the proposed MAD.  We can find that the baseline \texttt{GPT-3.5-Turbo} (even the more powerful \texttt{GPT-4}) incorrectly translates the source words literally. Because of the DoT issue, Self-Reflect cannot rectify the literal translation. The proposed MAD framework, which explores divergent chain-of-thoughts, can generate the free translation of the underlined words within the source sentences. 

\subsection{Results on Counter-Intuitive AR}
\label{subsec_ar}

\begin{table}[t!]
\centering
\begin{tabular}{l c}
\toprule
\bf Method  & \multicolumn{1}{c}{\bf ACC (\%)} \\
\midrule
\bf\texttt{GPT-4}         &  51.0 \\
\hline
\bf\texttt{GPT-3.5-Turbo} & 26.0 \\
\bf ~~+ CoT        & 28.0       \\
\bf ~~+ Self-Consistency      & 29.5   \\
\bf ~~+ Self-Reflect  &  27.5         \\ 
\hdashline
\bf ~~+ MAD               & \textbf{37.0} \\
\bottomrule
\end{tabular}
\vspace{-5pt}
\caption{Accuracy on Counter-Intuitive AR.}
\label{tab:CIAR}
\vspace{-10pt}
\end{table}

\paragraph{Results.} Table~\ref{tab:CIAR} lists the results in terms of reasoning accuracy.
We can observe that Self-Reflect only marginally improves over the baseline \texttt{GPT-3.5-Turbo}, while CoT and Self-Consistency bring more improvements.
Our MAD framework, though not as good as \texttt{GPT-4}, outperforms all the other compared methods based on \texttt{GPT-3.5-Turbo}, which further demonstrates its effectiveness. We also validate MAD on math and symbolic reasoning tasks and report our results in Appendix~\ref{app:math_symbolic_results}.

\paragraph{Case Study.} Figure~\ref{fig:method} shows an example on Counter-Intuitive AR. We find both CoT and Self-Reflect fail to reach the right answer by mistakenly outputing $3$. With divergent thinking, our MAD framework emerges ``\textit{we need to consider both the rotation around circle B and the rotation of circle A itself}'' and find the correct answer $4$.

\section{Analysis}
\label{sec:analysis}

In this section, we present a qualitative analysis to provide some insights how MAD works.
Unless otherwise stated, we report the overall results on the Common~MT dataset.

\begin{table}[t]
\centering
\begin{tabular}{l cc cc}
\toprule
\bf Method & \bf Bias$\downarrow$ & \bf Diversity$\uparrow$\\
\midrule
\textbf{Self-Reflect}  & 29.0 & 19.3\\
\textbf{MAD}                  & 24.8 & 49.7\\
\bottomrule
\end{tabular}
\vspace{-5pt}
\caption{Mitigation of Degeneration-of-Thought.}
\label{tab:mitigate-dot}
% \vspace{-5pt}
\end{table}

\subsection{Mitigation of DoT}

As mentioned in the Section~\ref{sec:introduction}, the DoT problem originates from three factors: (1) Bias and Distorted Perception, (2) Rigidity and Resistance to Change, and (3) Limited External Feedback. In our MAD framework, we introduce the views of other agents in the form of debates, solving the phenomenon of limited external feedback (problem 3). Next, this section will delve into the mitigation of problems 1 and 2 through experiments.

\begin{itemize}[leftmargin=10pt]
    \item \textbf{Bias}: We observe that LLMs often rely on direct intuition, which can lead to incorrect or inappropriate responses. To address this problem, we use human evaluation to determine the ambiguity error rate of LLMs' responses, examining if the LLM's output is biased.
    \item \textbf{Diversity}: LLMs are resistant to changing their answers and lack diverse reflection. The diversity of the translations is evaluated using the Self-BLEU score~\citep{yin2020meta}. In other words, methods lacking diverse reflection produce more similar translation candidates. Consequently, higher Self-BLEU scores mean lower diversity. We calculate text diversity via:
    \begin{equation}
    \scalebox{0.84}{$
    \mathrm{Diversity} = 100-\mathrm{Self\_BLEU}\left(Cand_1, Cand_2\right)
    $}
    \label{formula:diversity}
    \end{equation}
\end{itemize}

In formula~(\ref{formula:diversity}), candidates 1 and 2 represent the initial translation (base answer in Self-Reflection or affirmative side's response in MAD) and the current translation (possible modified answer after Self-Reflection or negative side's response in MAD).

As shown in Table~\ref{tab:mitigate-dot}, 
Bias and Rigidity are significant factors causing DoT. In addition, addressing these biases and stereotypes through self-reflection can be challenging. MAD framework effectively corrects inherent biases in translation, mitigates DoT, and considerably improves performance.

\subsection{Analysis of Judge}

In this section, we analyze the behavior of the judge for different settings of the debaters.

\paragraph{Strong debaters with a weak judge work better than the reverse.}
To understand the roles of debaters and judge in MAD, we employ various combinations of models to initialize the agents. Specifically, we utilize the smaller language model (\texttt{vicuna-13b-v1.5-16k}) as a judge to evaluate the debate results of the more powerful LLMs (\texttt{GPT-3.5-Turbo}), and vice versa. 

The detailed experimental findings are presented in Table~\ref{tab:weak-judge}. The quality of the debaters' responses significantly impact the performance ceiling of MAD. Regardless of the model chosen for the judge, Turbo debaters consistently generate superior translations compared to Vicuna. In addition, the selection of the judge agent plays a secondary role. When Turbo debaters are involved, Vicuna, serving as the judge, underperforms Turbo across all test sets.

\begin{table}[t]
\centering
\begin{tabular}{l rr}
\toprule
\bf Judge LLM & \bf COMET & \bf HUMAN \\
\midrule
\multicolumn{3}{c}{\texttt{Vicuna-13b as Debaters}} \\
{\bf \texttt{Vicuna-13b}}  & 79.9 & 3.20 \\
{\bf \texttt{GPT-3.5-Turbo}}   & 80.4 & 3.25 \\
\midrule
\multicolumn{3}{c}{\texttt{GPT-3.5-Turbo as Debaters}} \\
{\bf \texttt{Vicuna-13b}}  & 83.2 & 3.47 \\
{\bf \texttt{GPT-3.5-Turbo}}   & 84.4 & 3.69 \\
\bottomrule
\end{tabular}
\vspace{-5pt}
\caption{Translation performance with different judge.}
\label{tab:weak-judge}
% \vspace{-12pt}
\end{table}

\paragraph{LLM may not act as an impartial judge when different LLMs are used as debaters.}
% \paragraph{LLM-based judge has a preference for outputs generated by itself.}
We study the behavior of agents by calculating how many times the judge chooses the answers of each debater as the final solution in different scenarios. The results are listed in Table~\ref{tab:behavior-agent} and we have the following observations:
\begin{itemize}[leftmargin=10pt]
    \item \textit{Same  LLM for All Agents} (Rows ~\textcircled{\small{1}} and \textcircled{\small{2}}): We find that the judge consistently favors the negative side, which is believed to contribute to the performance improvement in MAD. When encountering complex tasks, the affirmative side tends to make mistakes that should be corrected by the opposing side to achieve improvements.
    \item \textit{Debaters of Different LLMs} (Rows~\textcircled{\small{3}} and \textcircled{\small{4}}): We find that the judge shows a preference to the side with the same LLM as the backbone. This bias indicates that
LLMs might not be a fair judge~\citep{wang2023FairEval} when different LLMs are used for the agents.
\end{itemize}

\subsection{Analysis of Debaters}

In this section, we will discuss several factors of debaters that would affect the performance of MAD: \textit{debater number}, \textit{debate level}, and \textit{debate iteration}.

\paragraph{Increasing the number of debaters fails when backbone LLMs are poor at long-text modeling.}
It seems intuitive that increasing the number of debaters would enhance diversity of thought and subsequently improve performance. However, as shown in Table~\ref{tab:more-debaters}, an increase in the number of debaters has resulted in varying degrees of performance reduction.

\begin{figure}[t]
    \centering
    \includegraphics[height=0.67\linewidth]{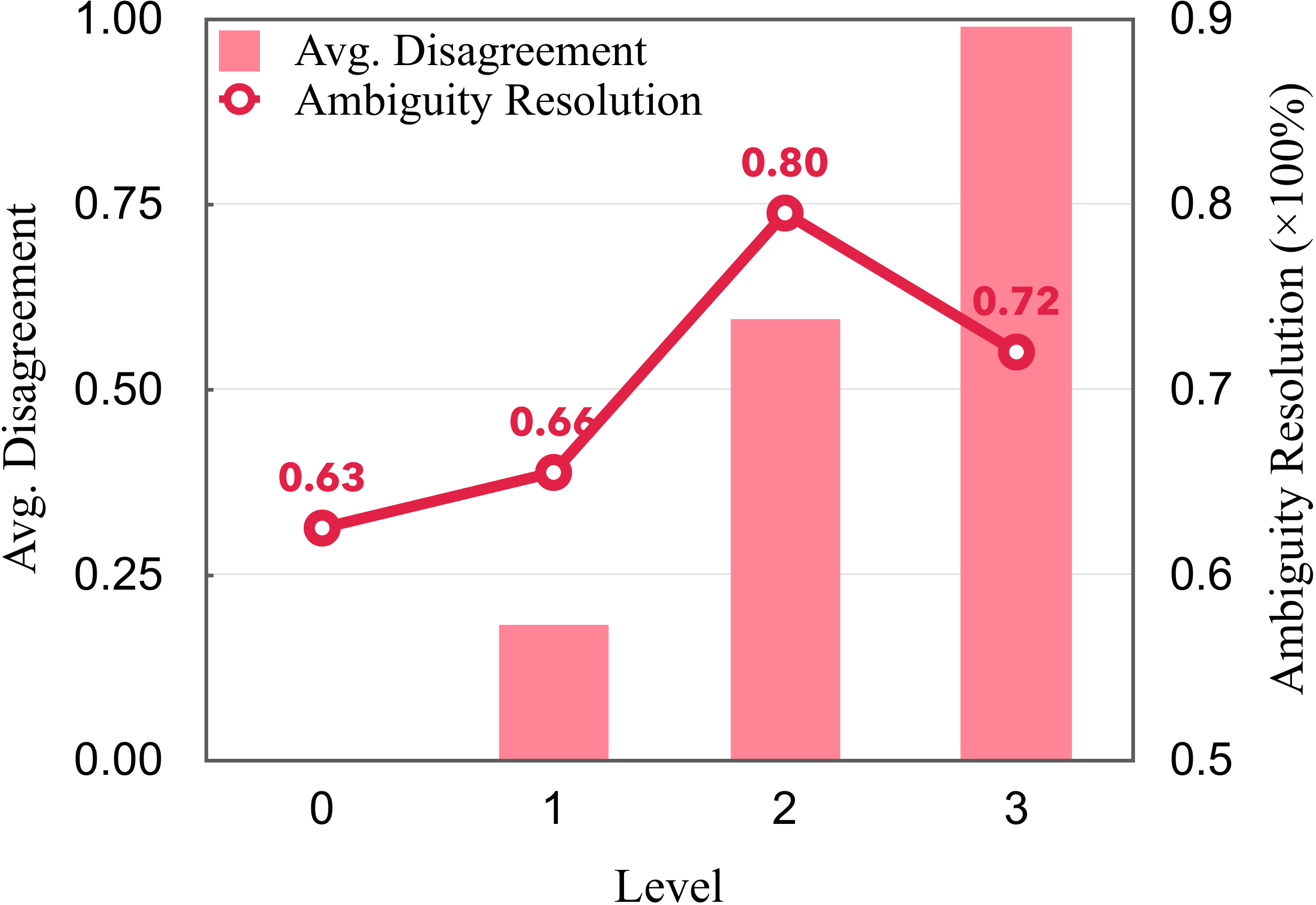}
    \vspace{-5pt}
    \caption{Translation performance with respect to the debate level on Lexical.}
    \label{fig:tit-for-tat}
    % \vspace{-5pt}
\end{figure}

\begin{table}[!t]
\centering
\setlength\tabcolsep{5pt}
\begin{tabular}{c ccc rrr}
\toprule
\multirow{2}{*}{\bf ID} & \multirow{2}{*}{\bf Jud} & \multicolumn{2}{c}{\bf Debater} & \multicolumn{3}{c}{\bf Winner}\\
\cmidrule(lr){3-4}\cmidrule(lr){5-7}
&   &  \bf Aff & \bf Neg &  \bf Aff & \bf Neg & \bf Tie  \\
\cmidrule(lr){1-4}\cmidrule(lr){5-7}
\textcircled{\small{1}} & \tt Turbo & \tt Turbo & \tt Turbo & 87 & \bf 104 & 9 \\
\textcircled{\small{2}} & \tt GPT-4 & \tt GPT-4 & \tt GPT-4 & 67 & \bf 124 & 9 \\
\hline
\textcircled{\small{3}} & \multirow{2}{*}{\tt GPT-4} & \tt Turbo & \tt GPT-4 & 52 & \bf 136 & 12 \\
\textcircled{\small{4}} & & \tt GPT-4 & \tt Turbo& \bf 120 & 77 & 3 \\
\bottomrule
\end{tabular}
\vspace{-5pt}
\caption{Number of times the judge chooses the answers of each debater based on different LLM.}
\label{tab:behavior-agent}
% \vspace{-10pt}
\end{table}

\begin{figure}
    \centering
    \includegraphics[height=0.67\linewidth]{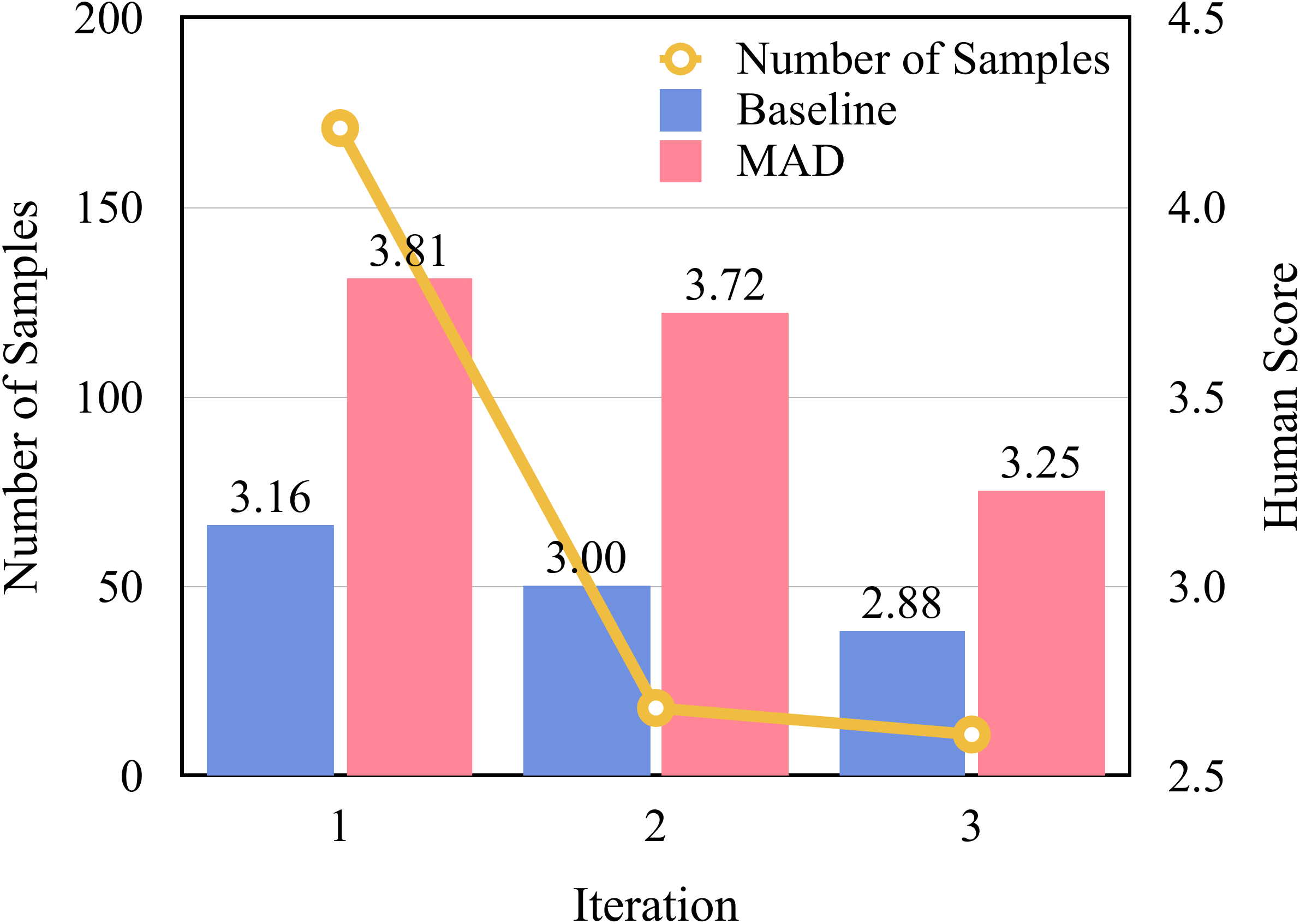}
    \vspace{-5pt}
    \caption{Distribution of iteration rounds and a human score of each iteration subset.}
    \label{fig:distribution}
\end{figure}

\begin{table}
\centering
\begin{tabular}{l rr}
\toprule
\bf \# of Debaters & \small \bf COMET & \small \bf HUMAN \\
\midrule
\bf 2 (Default)   & 84.4 & 3.69 \\
\hdashline
\bf 3  & 83.1 & 3.58 \\
\bf 4  & 82.9 & 3.49 \\
\bottomrule
\end{tabular}
\vspace{-5pt}
\caption{Translation performance with more debaters.}
\label{tab:more-debaters}
% \vspace{-10pt}
\end{table}

To address this issue, we manually analyze the debate processes in approximately 10\% of the test subset. As the number of debaters increases, the length and complexity of the text also increase. Such LLM-based debaters tend to forget the views of other debaters during the debate. Moreover, it becomes more challenging for the judge to extract information from the debates for summarization. This suggests that the key challenge of MAD with more debaters lies in the limitations of the LLMs to handle long texts~\citep{liu2023lost}.

\paragraph{Appropriate "tit for tat" is beneficial for effective debate.}
We then study how the intensity of ``tit for tat'' affects the performance of MAD.
To achieve so, we design different instructions~(see Table~\ref{tab:tit-for-tat-prompt} in Appendix) to initialize the debaters' meta prompt. As shown in Figure~\ref{fig:tit-for-tat}, asking the debaters to ``tit for tat''~(i.e., higher disagreement) is necessary for MAD to achieve good performance. However, we find that ``\textit{must disagree with each other on every point }'' (with a disagreement of 0.988) does not lead to the best performance.
We speculate that continuous disagreement without finding common ground can contribute to polarization, where the debate becomes more about winning the argument than seeking truth or understanding. This can reinforce pre-existing biases and make it difficult to reach a meaningful consensus.

\paragraph{Complex questions require more iteration rounds of debate.}
In our experimental setup, we did not implement any additional stopping strategies besides setting the maximum debate iteration to 3. In other words, the judge can take an \textit{adaptive break} if it believes the optimal answer has already been obtained, efficiently ending the debate early.

To understand the distribution of iteration rounds and factors contributing to a longer debate process, we analyze the experimental results and present them in Figure~\ref{fig:distribution}. In the majority of cases, the optimal answer can be achieved through a single round of debate, demonstrating the efficiency of MAD. However, when translating more complex sentences (subsets with lower human scores), the judge requires additional iterations to gather adequate information from the debaters before making a final decision.
We also find that our MAD framework consistently brings performance improvements across all the three subsets, demonstrating its effectiveness.

\begin{figure}
    \centering
    \includegraphics[height=0.67\linewidth]{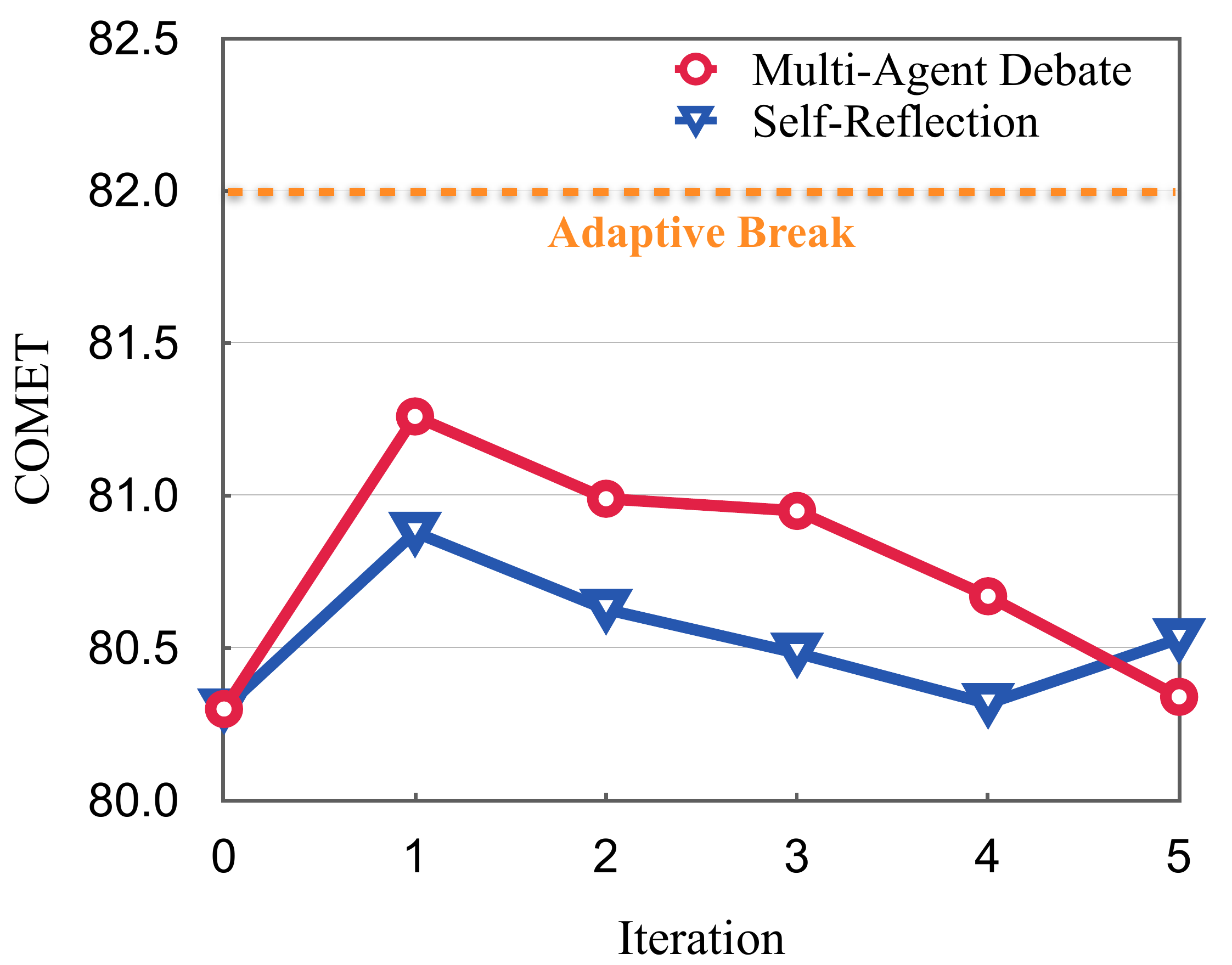}
    \vspace{-5pt}
    \caption{Performance with respect to the iteration of debate or self-reflection.}
    \label{fig:comet-iteration}
    % \vspace{-10pt}
\end{figure}

\paragraph{Adaptive break plays an important role to conclude the debate in the optimal moment.}
Intuitively, longer debates would encourage more diverse thinking. It raises the question of how the model's performance would be affected if constrained to conclude at a specific debate round.
For each iteration, we force the judge $J$ to extract the final answer ($a = J_e(H)$) instead of adaptively breaking the debate as in MAD. 

As shown in figure~\ref{fig:comet-iteration}, we can observe that MAD performs better than self-reflection as the iteration increases. However, the highest COMET score appears at the first iteration and is also lower than the result of the adaptive break. It indicates that, for most examples, MAD can generate good translations at the first iteration such that the debate should be stopped.
Forcing the debate to continue will harm the translation results, which demonstrates the reasonableness of our adaptive break strategy.

\section{Related Work}

\paragraph{Chain-of-Thought Prompting.}
Recently,~\cite{wei2022CoT} has proposed chain-of-thought (CoT) prompting to improve the reasoning ability of LLMs. Specifically, CoT prompts LLMs to generate a series of intermediate steps that lead to the final answer of a multi-step problem. Most earlier work primarily concentrates on two main aspects: prompt design and decoding strategies. Zero-shot CoT~\citep{kojimalarge} employs the trigger sentence ``Let’s think step by step'' to provide guidance for the decoding of LLMs. Advanced sampling strategies have been explored to improve CoT by generating diverse reasoning paths, e.g., Self-Consistency~\citep{wang2022self}, Auto-CoT~\citep{zhang2022automatic}, Active-Prompting~\citep{diao2023active}, Complexity-based Consistency~\citep{fu2022complexity}, Multi-Chain Reasoning~\citep{yoran2023answering}, and Progressive-Hint Prompting~\citep{zheng2023progressive}. 

With the emergence of powerful LLMs, approaches based on self-evaluation have attracted increasing attention. These approaches involve the generation of initial output, followed by evaluating the output to acquire feedback, which is then utilized to refine the output. Evaluation feedback can come from the model itself, e.g., Self-refine~\citep{madaan2023selfrefine} and Tree of Thoughts~\citep{yao2023tree}) or external environments, e.g., QAaP~\citep{zhu2023question} and Reflection~\citep{shinn2023reflexion}. The intuition behind these approaches involves the utilization of robust LLMs to mimic the human cognition process. 

\paragraph{Generative Agents.}
Recently, LLM-based multi-agent intelligent, e.g., Generative Agents~\citep{park2023generative}, Ghost in the Minecraft~\citep{zhu2023ghost}, GPT-Bargaining~\citep{fu2023improving}, has drawn significant attention for enabling simulations of human behavior. Our work follows this research line to address the DoT problem of LLMs. Concurrent with our work, a few studies~\citep{xiong2023diving,du2023improving} also explore the multi-agent debate framework to  enhance the reasoning ability of LLMs. The main differences between our MAD framework and these works are: 
(1) we introduce an additional judge with an adaptive break mechanism to decide the optimal moment to conclude the debate;
(2) our work aims to address the DoT problem, which is an inherent deficiency of LLMs; 
and (3) we empirically find that our MAD framework can yield enhanced performance by employing agents with the identical backbone LLM.

\section{Conclusion}

We propose and define the Degeneration-of-Thought (DoT) problem in self-reflection, and address it by proposing the Multi-Agent Debate~(MAD) framework to explore divergent chain-of-thoughts.
We demonstrate the effectiveness of MAD on two challenging tasks and find that \texttt{GPT-3.5-Turbo} with MAD can even surpass \texttt{GPT-4} on the Common~MT dataset.
Extensive analyses suggest that the adaptive break strategy of debate and the modest level of ``tit for tat'' state are required for MAD to obtain good performance. 
Complex samples require more rounds of debate.
More interestingly, we find that LLMs might not be a fair judge if different LLMs are used for agents. 

Future work includes scheduling more agents in the debate in an appropriate manner, multi-agent intelligence for board games, and AI feedback for model alignment.

\section*{Limitations}

A limitation of this work is that our method requires more time cost, as agents need to engage in multiple rounds of interaction to present and refute arguments. 
Moreover, current LLM-based agents may struggle to maintain coherence and relevance in long context scenarios, leading to potential misunderstandings and loss of context.
Enhancing long-text modeling capability of large language models remains a future challenge.
LLM-based judge may have a preference for outputs generated by itself. To mitigate this bias within the MAD framework, we recommend that all roles, including both the judge and debaters, utilize the same LLM, or alternatively, that the judge and debaters employ distinct LLMs.

\bibliography{anthology, custom}

\begin{thebibliography}{39}
\providecommand{\natexlab}[1]{#1}

\bibitem[{Bang et~al.(2023)Bang, Cahyawijaya, Lee, Dai, Su, Wilie, Lovenia, Ji, Yu, Chung et~al.}]{bang2023multitask}
Yejin Bang, Samuel Cahyawijaya, Nayeon Lee, Wenliang Dai, Dan Su, Bryan Wilie, Holy Lovenia, Ziwei Ji, Tiezheng Yu, Willy Chung, et~al. 2023.
\newblock A multitask, multilingual, multimodal evaluation of chatgpt on reasoning, hallucination, and interactivity.
\newblock In \emph{Proceedings of the 13th International Joint Conference on Natural Language Processing and the 3rd Conference of the Asia-Pacific Chapter of the Association for Computational Linguistics (Volume 1: Long Papers)}, pages 675--718.

\bibitem[{Bortolotti(2011)}]{bortolotti2011does}
Lisa Bortolotti. 2011.
\newblock Does reflection lead to wise choices?
\newblock \emph{Philosophical Explorations}, 14(3):297--313.

\bibitem[{Cobbe et~al.(2021)Cobbe, Kosaraju, Bavarian, Chen, Jun, Kaiser, Plappert, Tworek, Hilton, Nakano et~al.}]{cobbe2021training}
Karl Cobbe, Vineet Kosaraju, Mohammad Bavarian, Mark Chen, Heewoo Jun, Lukasz Kaiser, Matthias Plappert, Jerry Tworek, Jacob Hilton, Reiichiro Nakano, et~al. 2021.
\newblock Training verifiers to solve math word problems.
\newblock \emph{arXiv preprint arXiv:2110.14168}.

\bibitem[{Daniel(2017)}]{kahneman2011thinking}
Kahneman Daniel. 2017.
\newblock \emph{Thinking, fast and slow}.
\newblock Farrar, Straus and Giroux.

\bibitem[{Diao et~al.(2023)Diao, Wang, Lin, and Zhang}]{diao2023active}
Shizhe Diao, Pengcheng Wang, Yong Lin, and Tong Zhang. 2023.
\newblock Active prompting with chain-of-thought for large language models.
\newblock \emph{arXiv preprint arXiv:2302.12246}.

\bibitem[{Du et~al.(2023)Du, Li, Torralba, Tenenbaum, and Mordatch}]{du2023improving}
Yilun Du, Shuang Li, Antonio Torralba, Joshua~B Tenenbaum, and Igor Mordatch. 2023.
\newblock Improving factuality and reasoning in language models through multiagent debate.
\newblock \emph{arXiv preprint arXiv:2305.14325}.

\bibitem[{Fu et~al.(2023)Fu, Peng, Khot, and Lapata}]{fu2023improving}
Yao Fu, Hao Peng, Tushar Khot, and Mirella Lapata. 2023.
\newblock Improving language model negotiation with self-play and in-context learning from ai feedback.
\newblock \emph{arXiv preprint arXiv:2305.10142}.

\bibitem[{Fu et~al.(2022)Fu, Peng, Sabharwal, Clark, and Khot}]{fu2022complexity}
Yao Fu, Hao Peng, Ashish Sabharwal, Peter Clark, and Tushar Khot. 2022.
\newblock Complexity-based prompting for multi-step reasoning.
\newblock \emph{arXiv preprint arXiv:2210.00720}.

\bibitem[{Garcia et~al.(2023)Garcia, Bansal, Cherry, Foster, Krikun, Johnson, and Firat}]{garcia2023unreasonable}
Xavier Garcia, Yamini Bansal, Colin Cherry, George Foster, Maxim Krikun, Melvin Johnson, and Orhan Firat. 2023.
\newblock The unreasonable effectiveness of few-shot learning for machine translation.
\newblock In \emph{International Conference on Machine Learning}, pages 10867--10878. PMLR.

\bibitem[{Gou et~al.(2023)Gou, Shao, Gong, Shen, Yang, Duan, and Chen}]{gou2023critic}
Zhibin Gou, Zhihong Shao, Yeyun Gong, Yelong Shen, Yujiu Yang, Nan Duan, and Weizhu Chen. 2023.
\newblock Critic: Large language models can self-correct with tool-interactive critiquing.

\bibitem[{He et~al.(2020)He, Wang, Xiong, and Liu}]{he-etal-2020-box}
Jie He, Tao Wang, Deyi Xiong, and Qun Liu. 2020.
\newblock \href {https://doi.org/10.18653/v1/2020.findings-emnlp.327} {The box is in the pen: Evaluating commonsense reasoning in neural machine translation}.
\newblock In \emph{Findings of the Association for Computational Linguistics: EMNLP 2020}, pages 3662--3672, Online. Association for Computational Linguistics.

\bibitem[{He et~al.(2024)He, Liang, Jiao, Zhang, Yang, Wang, Tu, Shi, and Wang}]{he2023exploring}
Zhiwei He, Tian Liang, Wenxiang Jiao, Zhuosheng Zhang, Yujiu Yang, Rui Wang, Zhaopeng Tu, Shuming Shi, and Xing Wang. 2024.
\newblock Exploring human-like translation strategy with large language models.
\newblock \emph{Transactions of the Association for Computational Linguistics}, 12:229--246.

\bibitem[{Hendy et~al.(2023)Hendy, Abdelrehim, Sharaf, Raunak, Gabr, Matsushita, Kim, Afify, and Awadalla}]{gpt-mt-2023}
Amr Hendy, Mohamed Abdelrehim, Amr Sharaf, Vikas Raunak, Mohamed Gabr, Hitokazu Matsushita, Young~Jin Kim, Mohamed Afify, and Hany~Hassan Awadalla. 2023.
\newblock How good are gpt models at machine translation? a comprehensive evaluation.
\newblock \emph{arXiv preprint arXiv:2302.09210}.

\bibitem[{Hosseini et~al.(2014)Hosseini, Hajishirzi, Etzioni, and Kushman}]{hosseini2014learning}
Mohammad~Javad Hosseini, Hannaneh Hajishirzi, Oren Etzioni, and Nate Kushman. 2014.
\newblock Learning to solve arithmetic word problems with verb categorization.
\newblock In \emph{Proceedings of the 2014 Conference on Empirical Methods in Natural Language Processing (EMNLP)}, pages 523--533.

\bibitem[{Jiao et~al.(2023)Jiao, Wang, Huang, Wang, Shi, and Tu}]{jiao2023chatgpt}
Wenxiang Jiao, Wenxuan Wang, Jen-tse Huang, Xing Wang, Shuming Shi, and Zhaopeng Tu. 2023.
\newblock Is chatgpt a good translator? yes with gpt-4 as the engine.
\newblock \emph{arXiv preprint arXiv:2301.08745}.

\bibitem[{Keestra(2017)}]{keestra2017metacognition}
Machiel Keestra. 2017.
\newblock Metacognition and reflection by interdisciplinary experts: Insights from cognitive science and philosophy.
\newblock \emph{Issues in Interdisciplinary Studies}, 35:121--169.

\bibitem[{Kojima et~al.(2022)Kojima, Gu, Reid, Matsuo, and Iwasawa}]{kojimalarge}
Takeshi Kojima, Shixiang~Shane Gu, Machel Reid, Yutaka Matsuo, and Yusuke Iwasawa. 2022.
\newblock Large language models are zero-shot reasoners.
\newblock \emph{Advances in neural information processing systems}, 35:22199--22213.

\bibitem[{Kong et~al.(2022)Kong, Li, Zhang, Huang, and Wu}]{kong2022eliciting}
Yuqing Kong, Yunqi Li, Yubo Zhang, Zhihuan Huang, and Jinzhao Wu. 2022.
\newblock Eliciting thinking hierarchy without a prior.
\newblock \emph{Advances in Neural Information Processing Systems}, 35:13329--13341.

\bibitem[{Liu et~al.(2024)Liu, Lin, Hewitt, Paranjape, Bevilacqua, Petroni, and Liang}]{liu2023lost}
Nelson~F Liu, Kevin Lin, John Hewitt, Ashwin Paranjape, Michele Bevilacqua, Fabio Petroni, and Percy Liang. 2024.
\newblock Lost in the middle: How language models use long contexts.
\newblock \emph{Transactions of the Association for Computational Linguistics}, 12:157--173.

\bibitem[{Madaan et~al.(2024)Madaan, Tandon, Gupta, Hallinan, Gao, Wiegreffe, Alon, Dziri, Prabhumoye, Yang et~al.}]{madaan2023selfrefine}
Aman Madaan, Niket Tandon, Prakhar Gupta, Skyler Hallinan, Luyu Gao, Sarah Wiegreffe, Uri Alon, Nouha Dziri, Shrimai Prabhumoye, Yiming Yang, et~al. 2024.
\newblock Self-refine: Iterative refinement with self-feedback.
\newblock \emph{Advances in Neural Information Processing Systems}, 36.

\bibitem[{Park et~al.(2023)Park, O'Brien, Cai, Morris, Liang, and Bernstein}]{park2023generative}
Joon~Sung Park, Joseph O'Brien, Carrie~Jun Cai, Meredith~Ringel Morris, Percy Liang, and Michael~S Bernstein. 2023.
\newblock Generative agents: Interactive simulacra of human behavior.
\newblock In \emph{Proceedings of the 36th Annual ACM Symposium on User Interface Software and Technology}, pages 1--22.

\bibitem[{Pilault et~al.(2023)Pilault, Garcia, Bra{\v{z}}inskas, and Firat}]{pilault2023interactive}
Jonathan Pilault, Xavier Garcia, Arthur Bra{\v{z}}inskas, and Orhan Firat. 2023.
\newblock Interactive-chain-prompting: Ambiguity resolution for crosslingual conditional generation with interaction.
\newblock In \emph{Proceedings of the 13th International Joint Conference on Natural Language Processing and the 3rd Conference of the Asia-Pacific Chapter of the Association for Computational Linguistics (Volume 1: Long Papers)}, pages 455--483.

\bibitem[{Roy and Roth(2015)}]{roy2015solving}
Subhro Roy and Dan Roth. 2015.
\newblock Solving general arithmetic word problems.
\newblock In \emph{Proceedings of the 2015 Conference on Empirical Methods in Natural Language Processing}, pages 1743--1752.

\bibitem[{Shinn et~al.(2024)Shinn, Cassano, Gopinath, Narasimhan, and Yao}]{shinn2023reflexion}
Noah Shinn, Federico Cassano, Ashwin Gopinath, Karthik Narasimhan, and Shunyu Yao. 2024.
\newblock Reflexion: Language agents with verbal reinforcement learning.
\newblock \emph{Advances in Neural Information Processing Systems}, 36.

\bibitem[{Srivastava et~al.(2023)Srivastava, Rastogi, Rao, Shoeb, Abid, Fisch, Brown, Santoro, Gupta, Garriga-Alonso et~al.}]{srivastava2023beyond}
Aarohi Srivastava, Abhinav Rastogi, Abhishek Rao, Abu Awal~Md Shoeb, Abubakar Abid, Adam Fisch, Adam~R Brown, Adam Santoro, Aditya Gupta, Adri{\`a} Garriga-Alonso, et~al. 2023.
\newblock Beyond the imitation game: Quantifying and extrapolating the capabilities of language models.
\newblock \emph{Transactions on Machine Learning Research}.

\bibitem[{Suzgun et~al.(2023)Suzgun, Scales, Sch{\"a}rli, Gehrmann, Tay, Chung, Chowdhery, Le, Chi, Zhou et~al.}]{suzgun2023challenging}
Mirac Suzgun, Nathan Scales, Nathanael Sch{\"a}rli, Sebastian Gehrmann, Yi~Tay, Hyung~Won Chung, Aakanksha Chowdhery, Quoc Le, Ed~Chi, Denny Zhou, et~al. 2023.
\newblock Challenging big-bench tasks and whether chain-of-thought can solve them.
\newblock In \emph{Findings of the Association for Computational Linguistics: ACL 2023}, pages 13003--13051.

\bibitem[{Wang et~al.(2023)Wang, Li, Chen, Cai, Zhu, Lin, Cao, Liu, Liu, and Sui}]{wang2023FairEval}
Peiyi Wang, Lei Li, Liang Chen, Zefan Cai, Dawei Zhu, Binghuai Lin, Yunbo Cao, Qi~Liu, Tianyu Liu, and Zhifang Sui. 2023.
\newblock Large language models are not fair evaluators.
\newblock \emph{arXiv preprint arXiv:2305.17926}.

\bibitem[{Wang et~al.(2022)Wang, Wei, Schuurmans, Le, Chi, Narang, Chowdhery, and Zhou}]{wang2022self}
Xuezhi Wang, Jason Wei, Dale Schuurmans, Quoc Le, Ed~Chi, Sharan Narang, Aakanksha Chowdhery, and Denny Zhou. 2022.
\newblock Self-consistency improves chain of thought reasoning in language models.
\newblock \emph{arXiv preprint arXiv:2203.11171}.

\bibitem[{Wei et~al.(2022)Wei, Wang, Schuurmans, Bosma, Xia, Chi, Le, Zhou et~al.}]{wei2022CoT}
Jason Wei, Xuezhi Wang, Dale Schuurmans, Maarten Bosma, Fei Xia, Ed~Chi, Quoc~V Le, Denny Zhou, et~al. 2022.
\newblock Chain-of-thought prompting elicits reasoning in large language models.
\newblock \emph{Advances in neural information processing systems}, 35:24824--24837.

\bibitem[{Wu et~al.(2023)Wu, Wang, Wan, Jiao, and Lyu}]{wu2023chatgpt4gec}
Haoran Wu, Wenxuan Wang, Yuxuan Wan, Wenxiang Jiao, and Michael Lyu. 2023.
\newblock Chatgpt or grammarly? evaluating chatgpt on grammatical error correction benchmark.
\newblock \emph{arXiv preprint arXiv:2303.13648}.

\bibitem[{Xiong et~al.(2023)Xiong, Ding, Cao, Liu, and Qin}]{xiong2023diving}
Kai Xiong, Xiao Ding, Yixin Cao, Ting Liu, and Bing Qin. 2023.
\newblock Diving into the inter-consistency of large language models: An insightful analysis through debate.
\newblock \emph{arXiv preprint arXiv:2305.11595}.

\bibitem[{Yao et~al.(2024)Yao, Yu, Zhao, Shafran, Griffiths, Cao, and Narasimhan}]{yao2023tree}
Shunyu Yao, Dian Yu, Jeffrey Zhao, Izhak Shafran, Tom Griffiths, Yuan Cao, and Karthik Narasimhan. 2024.
\newblock Tree of thoughts: Deliberate problem solving with large language models.
\newblock \emph{Advances in Neural Information Processing Systems}, 36.

\bibitem[{Yin et~al.(2020)Yin, Li, Li, and Li}]{yin2020meta}
Haiyan Yin, Dingcheng Li, Xu~Li, and Ping Li. 2020.
\newblock Meta-cotgan: A meta cooperative training paradigm for improving adversarial text generation.
\newblock In \emph{Proceedings of the AAAI Conference on Artificial Intelligence}, volume~34, pages 9466--9473.

\bibitem[{Yoran et~al.(2023)Yoran, Wolfson, Bogin, Katz, Deutch, and Berant}]{yoran2023answering}
Ori Yoran, Tomer Wolfson, Ben Bogin, Uri Katz, Daniel Deutch, and Jonathan Berant. 2023.
\newblock Answering questions by meta-reasoning over multiple chains of thought.
\newblock In \emph{Proceedings of the 2023 Conference on Empirical Methods in Natural Language Processing}, pages 5942--5966.

\bibitem[{Zhang et~al.(2022)Zhang, Zhang, Li, and Smola}]{zhang2022automatic}
Zhuosheng Zhang, Aston Zhang, Mu~Li, and Alex Smola. 2022.
\newblock Automatic chain of thought prompting in large language models.
\newblock \emph{arXiv preprint arXiv:2210.03493}.

\bibitem[{Zheng et~al.(2023)Zheng, Liu, Xie, Li, and Li}]{zheng2023progressive}
Chuanyang Zheng, Zhengying Liu, Enze Xie, Zhenguo Li, and Yu~Li. 2023.
\newblock Progressive-hint prompting improves reasoning in large language models.
\newblock \emph{arXiv preprint arXiv:2304.09797}.

\bibitem[{Zhu et~al.(2023{\natexlab{a}})Zhu, Wang, Zhang, Zhang, Huang, Zhang, Yang et~al.}]{zhu2022core}
Xinyu Zhu, Junjie Wang, Lin Zhang, Yuxiang Zhang, Yongfeng Huang, Jiaxing Zhang, Yujiu Yang, et~al. 2023{\natexlab{a}}.
\newblock Solving math word problems via cooperative reasoning induced language models.
\newblock In \emph{The 61st Annual Meeting Of The Association For Computational Linguistics}.

\bibitem[{Zhu et~al.(2023{\natexlab{b}})Zhu, Yang, Chen, Li, Lou, and Yang}]{zhu2023question}
Xinyu Zhu, Cheng Yang, Bei Chen, Siheng Li, Jian-Guang Lou, and Yujiu Yang. 2023{\natexlab{b}}.
\newblock Question answering as programming for solving time-sensitive questions.
\newblock In \emph{Proceedings of the 2023 Conference on Empirical Methods in Natural Language Processing}, pages 12775--12790.

\bibitem[{Zhu et~al.(2023{\natexlab{c}})Zhu, Chen, Tian, Tao, Su, Yang, Huang, Li, Lu, Wang et~al.}]{zhu2023ghost}
Xizhou Zhu, Yuntao Chen, Hao Tian, Chenxin Tao, Weijie Su, Chenyu Yang, Gao Huang, Bin Li, Lewei Lu, Xiaogang Wang, et~al. 2023{\natexlab{c}}.
\newblock Ghost in the minecraft: Generally capable agents for open-world enviroments via large language models with text-based knowledge and memory.
\newblock \emph{arXiv preprint arXiv:2305.17144}.

\end{thebibliography}
% \bibliographystyle{acl_natbib}

% \clearpage

\appendix
\onecolumn
\newpage
\appendix

\section{Challenging Testbeds}
\label{app:testbeds}

We conduct experiments on two challenging tasks, namely, commonsense machine translation (i.e., Common~MT), and counter-intuitive arithmetic reasoning (i.e., Counter-Intuitive AR), which require deep levels of contemplation for LLMs. 

\subsection{Commonsense Machine Translation}

\begin{table*}[!h]
\small
\centering
\begin{tabular}{l p{0.19\columnwidth} p{0.27\columnwidth} p{0.25\columnwidth}}
\toprule
\bf Ambiguity Type & \bf Source Sentence & \bf Correct Reference & \bf Incorrect Translation \\
\midrule
\multirow{2}{*}{\bf Lexical} & \zh{\underline{吃}掉敌人一个师。} & \textcolor{blue}{Destroy} a division of the enemy. & \textcolor{red}{Eat up} an enemy division.\\
\cmidrule(lr){2-4}
& \zh{他喜欢\underline{吃}苹果。}& He likes to \textcolor{blue}{eat} apples.& He likes to \textcolor{red}{destory} apples.\\
\midrule
\multirow{4}{*}{\bf Contextless} & \zh{\underline{正在手术}的是健康的医生。} & A healthy doctor \textcolor{blue}{is doing surgery}. & What \textcolor{red}{is undergoing surgery} is a doctor who is healthy.\\
\cmidrule(lr){2-4}
& \zh{\underline{正在手术}的是生命垂危的病人。}& What \textcolor{blue}{is undergoing surgery} is a patient whose life is dying.& A patient whose life is dying \textcolor{red}{is doing surgery}. \\
\midrule
\multirow{4}{*}{\bf Contextual} & \zh{当地震袭击中国时，\underline{援助的是中国}。} & When the earthquake hit China, \textcolor{blue}{China was aided}. & When the earthquake hit China, \textcolor{red}{China has assisted}.\\
\cmidrule(lr){2-4}
& \zh{当地震袭击日本时，\underline{援助的是中国}。}& When the earthquake hit Japan, \textcolor{blue}{China has assisted}.& When the earthquake hit Japan, \textcolor{red}{China was aided}. \\
\bottomrule
\end{tabular}
\caption{Examples of lexical, contextual and contextless syntactic ambiguity from the Common~MT dataset. The underlined Chinese words are translated into the corresponding colored words in English. Best viewed in color.}
\label{tab:mt-example}
\end{table*}

The Common~MT dataset is composed of Chinese$\Rightarrow$English translation examples~\citep{he-etal-2020-box}, which are used to examine three types of ambiguity resolution abilities of translation models. Specifically, The Common~MT test set we used covers 200 examples of lexical ambiguity, 450 examples of contextless syntactic ambiguity, and 350 examples of contextual syntactic ambiguity. Within the challenging part of Common~MT, the authentic translation of each source sentence requires a proper understanding of common sense knowledge. While these ambiguous sentences might appear to have a straightforward translation, such a literal interpretation is erroneous. Failure to identify and address such ambiguities may result in inaccurate translations.

Table~\ref{tab:mt-example} lists some examples of these three types of ambiguity. Lexical ambiguity refers to words with multiple meanings in different contexts. Contextless and contextual syntactic ambiguity involve sentences with multiple interpretations, which can be resolved by context or common sense. As the lexical ambiguity of ``\zh{吃掉敌人一个师}'' shows, the source word ``\zh{吃掉}'' should be translated to ``destroy'' rather than the straightforward translation ``eat up'' by considering the common sense in the real world.

\subsection{Counter-Intuitive Arithmetic Reasoning}
Previous studies on thinking hierarchy~\citep{kahneman2011thinking} suggest that we humans have a fast and intuitive system and a slow and logical system, and tend to run the lower level system before the higher level one. Inspired by this, we created a more challenging dataset named Counter-Intuitive Arithmetic Reasoning (CIAR) to evaluate the reasoning abilities of LLMs at deep levels. 

\begin{table*}[!tb]
\centering
\begin{tabular}{l p{0.75\columnwidth}}
\toprule
\bf Components & \multicolumn{1}{c}{\bf Content} \\
\midrule
\bf Question & When Alice walks up the hill, her speed is 1 m/s and when she goes down the hill, her speed is 3 m/s. Then when Alice walks up and down the hill, what is her average speed?\\
\hline
\bf Correct Answer &  1.5 m/s\\
\hdashline
\bf Explanation & If Alice covers a distance of d going up and down the hill, then her total distance is 2d. Her time going up the hill is d/1 = d, and her time going down the hill is d/3. So, her total time is d + d/3 = 4d/3. Therefore, her average speed is 2d / (4d/3) = 3/2 m/s. \\
\hline
\bf Incorrect Answer & 2 m/s \\
\hdashline
\bf Explanation &  Alice's average speed can be calculated by adding her speed going up the hill and her speed going down the hill, and then dividing by 2. So, (1 m/s + 3 m/s) / 2 = 2 m/s. Therefore, Alice's average speed is 2 m/s. \\
\bottomrule
\end{tabular}
\caption{An example in Counter-Intuitive AR dataset.}
\label{tab:qa-example}
\end{table*}

\paragraph{Dataset Description.}
Our Counter-Intuitive AR dataset contains 200 questions collected from elicitation 
questions~\citep{kong2022eliciting}\footnote{https://elicitation.info/questionnaire/1/}, 
web data\footnote{https://www.geeksforgeeks.org/puzzles/} and additional manual derivatives of these questions.
Compared to the commonly-used datasets, e.g., MultiArith~\citep{roy2015solving}, GSM8K~\citep{cobbe2021training},  our dataset presents two distinct challenges:

\begin{itemize}[leftmargin=10pt]
    \item \textbf{Resistance to Intuition.} 
    The questions in our dataset are embedded in hidden traps designed to elicit intuitive and appealing answers that are often incorrect. This feature evaluates the abilities of LLMs to resist the traps of superficial expressions.
    \item \textbf{Multi-Step Reasoning.} 
    Each correct answer within the dataset requires a rigorous multi-step reasoning process, thereby evaluating the capacity of LLMs to engage in complex decision-making and problem-solving.
\end{itemize}

\paragraph{Dataset Format.}
In our Counter-Intuitive AR dataset, each example contains three key components (see Table~\ref{tab:qa-example} for an example). We elaborate on the details below:
\begin{itemize}[leftmargin=10pt]
    \item \textbf{Questions.} The questions in our dataset are designed to stimulate counter-intuitive thinking, which aims to challenge conventional decision-making by presenting situations where the immediate, intuitive response is often incorrect.  
    \item \textbf{Answers.} Each question is provided with a correct answer, which requires deep comprehension of the question and commonsense knowledge. Additionally, we also provide a plausible yet incorrect answer for comparison. 
    \item \textbf{Explanations.} We offer comprehensive explanations for each correct answer, detailing the step-by-step rationale that leads to the right solution. We also provide the seemingly logical reasoning process behind incorrect answers. This reasoning process highlights the potential pitfalls and misconceptions during decision-making, especially when intuition is prioritized over rigorous logical reasoning. 
    
\end{itemize}

\paragraph{Experimental Settings.}
During our experiments, we did not utilize the explanations from the dataset. We provided detailed explanations to facilitate subsequent researchers to understand how the correct answer was derived.

\section{Human Evaluation Details}
We implement human evaluation as follows:
\begin{itemize}[leftmargin=10pt]
    \item \textbf{Human Score}: We randomly shuffled the display order of the translated sentences from all methods in an anonymous manner. Then, employed three professional human translators (Krippendorff's Alpha = 0.76) to directly assess all methods together. Finally, we calculated the average scores for each methods.
    \item \textbf{Bias}: We also focus on whether the translation of specific words in CommonMT conforms to commonsense. Table~\ref{tab:mt-example} lists an example of lexical ambiguity, where the source word ``\zh{吃掉}'' should be translated to “destroy” rather than the straightforward translation “eat up”. Here, we asked the annotators to label each sentence as 1 (not conforming to commonsense) or 0 (conforming to commonsense), and report the degree of bias for the whole test set.
\end{itemize}

\clearpage

\section{Results on math and symbolic reasoning tasks}
\label{app:math_symbolic_results}

\begin{table}[!ht]
\centering
\begin{tabular}{l cc ccc}
\toprule
\multirow{2}{*}{\bf Method} & \multicolumn{2}{c}{\bf Math Reasoning	} & \multicolumn{3}{c}{\bf Symbolic Reasoning (BBH)}\\
\cmidrule(lr){2-3}\cmidrule(lr){4-6}
& \bf GSM & \bf AddSub & \bf Penguin & \bf Date & \bf Colored Objects \\
\midrule
\bf CoT          & 70.2 & 87.3 & 58.9 & 56.4 & 57.2 \\
\bf Self-Reflect & 70.8 & 87.6 & 61.0 & 58.0 & 58.0 \\
\bf MAD          & 73.8 & 92.1 & 63.7 & 65.2 & 58.8 \\
\bottomrule
\end{tabular}
% \vspace{-5pt}
\caption{Comparison of accuracy on math (e.g. GSM~\citep{cobbe2021training} and AddSub~\citep{hosseini2014learning}) and symbolic reasoning (three datasets from Big-Bench~\citep{srivastava2023beyond,suzgun2023challenging}).}
\label{tab:math_symbolic_results}
% \vspace{-10pt}
\end{table}

\section{Prompts for Different Debate Levels}

\begin{table*}[!h]
\centering
\begin{tabular}{l p{0.8\columnwidth}}
\toprule
\bf Level & \multicolumn{1}{c}{\bf Prompt} \\
\midrule
 \multirow{2}{*}{~0} & Both sides must reach a full consensus on every point of the debate. Every statement must be agreed upon by both sides.\\[2pt]
 \hline
\vspace{-1mm}
 \multirow{2}{*}{1} & Most of the debate should be characterized by disagreements, but there may still be a small amount of consensus on less significant points. \\[2pt]
 \hline
\vspace{-1mm}
 \multirow{2}{*}{2~(Default)} & It's not necessary to fully agree with each other's perspectives, as our objective is to find  the correct answer.\\[2pt]
 \hline
\vspace{-1mm}
 \multirow{2}{*}{3} & Both sides must disagree with each other on every point of the debate. There should be no consensus whatsoever.\\
\bottomrule
\end{tabular}
\caption{Prompts for different levels of ``tit for tat'' state. We modulate the level of ``tit for tat'' state outlined in Section~\ref{sec:method} through appending natural language instructions to the debaters' meta prompt.}
\label{tab:tit-for-tat-prompt}
\end{table*}

\section{Extra Computational Cost}

\begin{table}[!ht]
\centering
\begin{tabular}{l c}
\toprule
\bf Method & \bf Cost (\# Generated Tokens) \\
\midrule
\bf COT & 1.0 \\
\bf Self-Reflect   & 1.83$\times$ \\
\bf MAD            & 2.46$\times$ \\
\bottomrule
\end{tabular}
% \vspace{-5pt}
\caption{Extra generated tokens on the counter-intuitive AR task.}
\label{tab:computational_cost}
% \vspace{-10pt}
\end{table}

We measure the inference cost by the number of generated tokens. As listed in Table~\ref{tab:computational_cost}, the self-reflection cost is 1.83 times that of the baseline COT method. Meanwhile, our MAD approach incurs an additional 0.63 times the inference cost. This increase in cost is deemed acceptable given the significant benefit our approach offers in enhancing LLMs' divergent thinking capabilities without requiring external signals (interaction with humans).

\newpage

\section{Debate Process}

\subsection{Commonsense Machine Translation}

Table ~\ref{table:debate_case_translation} presents the debate process of example translation discussed in Section~\ref{subsec_mt}. 
    In this debate, both sides disputed the English translation of the Chinese sentence ``\zh{吃掉敌人一个师}''. The affirmative side proposed ``eat up an enemy division'', advocating for a more literal translation to depict the consumption or devouring of the enemy. In contrast, the negative side suggested ``eliminate an enemy division'', arguing this expression better conveys the destruction of the enemy in a military context. After the debate, the judge favored ``eliminate'', as it accurately reflected the severity and intent of the original military language. This debate highlights the complexities of translation, where context and nuance significantly impact the selection of appropriate words. In response to this issue, our MAD method helps LLMs understand semantics correctly and generate highly authentic translations.

\begin{table*}[!ht]
\centering
\vspace{10pt}
\begin{tabular}{p{15cm}}
\toprule
$<$\textbf{Debate Topic}$>$: \textit{What is the correct English translation of the following text: \zh{吃掉敌人一个师}?}\\
\vspace{-1mm}
\begin{minipage}[b]{0.03\columnwidth}
    \centering
    \raisebox{-.3\height}{\includegraphics[width=\linewidth]{imgs/emojis/devil.png}}
\end{minipage}
I believe the correct translation of the Chinese text ``\zh{吃掉敌人一个师}'' is  ``Eat up an enemy division.''  The word ``\zh{吃掉}'' literally means ``eat up'' or ``consume.'' \\
\vspace{-1mm}
\begin{minipage}[b]{0.03\columnwidth}
    \centering
    \raisebox{-.3\height}{\includegraphics[width=\linewidth]{imgs/emojis/angel.png}}
\end{minipage}
I disagree with your translation. I believe the correct translation of the Chinese text
``\zh{吃掉敌人一个师}'' is ``Eliminate an enemy division.'' While ``\zh{吃掉}'' can be translated as ``eat up'' or ``consume'', it can also mean ``eliminate'' or ``annihilate'' in a military context. In this case, it is more appropriate to use ``eliminate'' as it conveys the idea of completely destroying the enemy division.\\
\vspace{-1mm}
\begin{minipage}[b]{0.03\columnwidth}
    \centering
    \raisebox{-.3\height}{\includegraphics[width=\linewidth]{imgs/emojis/judge.png}}
\end{minipage}
Both sides presented valid arguments and translations that accurately convey the meaning of the original Chinese text. The choice between ``eat up'' and ``eliminate'' may depend on the context and tone of the text. Therefore, there is no clear preference for either side at this time.\\
\vspace{-1mm}
\begin{minipage}[b]{0.03\columnwidth}
    \centering
    \raisebox{-.3\height}{\includegraphics[width=\linewidth]{imgs/emojis/devil.png}}
\end{minipage}
While ``eliminate'' does convey the idea of completely destroying the enemy division, it doesn't capture the connotation of consuming or devouring the enemy that is implied by ``\zh{吃掉}''.\\
\vspace{-1mm}
\begin{minipage}[b]{0.03\columnwidth}
    \centering
    \raisebox{-.3\height}{\includegraphics[width=\linewidth]{imgs/emojis/angel.png}}
\end{minipage}
In a military context, ``\zh{吃掉}'' is often used to describe the complete destruction of the enemy, rather than the consumption of their resources or territory.\\
\vspace{-1mm}
\begin{minipage}[b]{0.03\columnwidth}
    \centering
    \raisebox{-.3\height}{\includegraphics[width=\linewidth]{imgs/emojis/judge.png}}
\end{minipage}
Both sides presented valid arguments, but the negative side's argument that ``eliminate'' better captures the intensity and seriousness of the military language used in the original Chinese text was more convincing. Additionally, ``eliminate'' accurately conveys the idea of completely destroying the enemy division. Therefore, the correct English translation of the Chinese text ``\zh{吃掉敌人一个师}'' is ``Eliminate an enemy division.''\\
\bottomrule
\end{tabular}
\caption{The debate process of translation example on Chinese sentence ``\zh{吃掉敌人一个师}''.}
\label{table:debate_case_translation}
\normalsize
\end{table*}

\newpage

\subsection{Counter-Intuitive Arithmetic Reasoning}

Table~\ref{table:debate_case_qa} shows the debate process of example answer in Table~\ref{tab:qa-example}. This debate is about calculating Alice's average speed while walking up and down a hill, with speeds given as 1 m/s uphill and 3 m/s downhill. 
Initially, it was incorrectly proposed by affirmative side to just average the uphill and downhill speeds. The negative side, which is correct, states that average speed is total distance divided by total time.
Given that the distances up and down the hill are equal, we must calculate the total time taken for the two directions before finding the average speed.
Therefore, the average speed is calculated using the sum of the uphill and downhill distances (twice the distance of the hill) and the sum of the time taken for each (d/1 for uphill and d/3 for downhill). This leads to an average speed of 1.5 m/s, not 2 m/s as the affirmative side proposed. And the judge also supports the response from the negative side. Here, LLMs made a common mistake initially in understanding the concept of average speed. Our MAD method encourages divergent thinking and effectively prevents intuitive but incorrect responses, leading to the correct method for calculating average speed.

\begin{table*}[!ht]
\centering
\vspace{10pt}
\begin{tabular}{p{15cm}}
\toprule
$<$\textbf{Debate Topic}$>$: \textit{When Alice walks up the hill, her speed is 1 m/s and when she goes down the hill, her speed is 3 m/s. Then when Alice walks up and down the hill, what is her average speed?}\\
\vspace{-1mm}
\begin{minipage}[b]{0.03\columnwidth}
    \centering
    \raisebox{-.3\height}{\includegraphics[width=\linewidth]{imgs/emojis/devil.png}}
\end{minipage}
Alice's average speed can be calculated by adding her speed going up the hill and her speed going down the hill, and then dividing by 2. So, (1 m/s + 3 m/s) / 2 = 2 m/s. Therefore, Alice's average speed is 2 m/s. \\
\vspace{-1mm}
\begin{minipage}[b]{0.03\columnwidth}
    \centering
    \raisebox{-.3\height}{\includegraphics[width=\linewidth]{imgs/emojis/angel.png}}
\end{minipage}
I disagree with your answer. The problem doesn't give us any information about the distance Alice covers while going up and down the hill. So, we cannot simply add her speeds and divide by 2 to get her average speed. We need to know the distance she covers to calculate her average speed.\\
\vspace{-1mm}
\begin{minipage}[b]{0.03\columnwidth}
    \centering
    \raisebox{-.3\height}{\includegraphics[width=\linewidth]{imgs/emojis/judge.png}}
\end{minipage}
The average speed is calculated by dividing the total distance by the total time taken. As Alice walks up and down the hill, the distance covered in both directions is the same. Therefore, the total distance covered is 2 times the distance of either direction. \\
\vspace{-1mm}
\begin{minipage}[b]{0.03\columnwidth}
    \centering
    \raisebox{-.3\height}{\includegraphics[width=\linewidth]{imgs/emojis/devil.png}}
\end{minipage}
That's a good point. But we can assume that Alice covers the same distance going up and down the hill. In that case, we can use the formula for average speed, which is total distance divided by total time. If Alice covers a distance of d going up and down the hill, then her total distance is 2d. Her time going up the hill is d/1 = d, and her time going down the hill is d/3. So, her total time is d + d/3 = 4d/3. Therefore, her average speed is 2d/(4d/3) = 3/2 m/s. \\
\vspace{-1mm}
\begin{minipage}[b]{0.03\columnwidth}
    \centering
    \raisebox{-.3\height}{\includegraphics[width=\linewidth]{imgs/emojis/angel.png}}
\end{minipage}
Thank you for providing your answer. This average speed of 1.5 m/s takes into account both the uphill and downhill speeds, providing a balanced representation of Alice's overall performance while walking up and down the hill. \\
\vspace{-1mm}
\begin{minipage}[b]{0.03\columnwidth}
    \centering
    \raisebox{-.3\height}{\includegraphics[width=\linewidth]{imgs/emojis/judge.png}}
\end{minipage}
Yes. Let's assume the distance is d. The time taken to walk up the hill is d/1 and the time taken to walk down the hill is d/3. The total time taken is d/1 + d/3 = 4d/3. Therefore, the average speed is (2d)/(4d/3) = 1.5 m/s. Hence, the negative side is correct. \\
\bottomrule
\end{tabular}
\caption{The debate process of arithmetic reasoning example.}
\label{table:debate_case_qa}
\end{table*}

\end{document}